\documentclass[11pt,letterpaper]{article}
\usepackage[T1]{fontenc}
\usepackage[sc]{mathpazo}
\usepackage{amsmath}
\usepackage[dvipsnames]{xcolor}
\usepackage{amssymb}
\usepackage{fullpage}
\usepackage[shortlabels,inline]{enumitem}
\usepackage{hyperref}
\usepackage{authblk}
\definecolor{pku-red}{RGB}{139,0,18}
\hypersetup{
    colorlinks=true,
    linkcolor=pku-red,      
    urlcolor=violet,
    citecolor=BlueViolet,
    pdffitwindow=true,
}

\title{How Large Language Models Need Symbolism}

\author{
  Xiaotie Deng\quad Hanyu Li
}

\affil{CFCS, School of Computer Science, Peking University\\
\texttt{\{xiaotie,lhydave\}@pku.edu.cn}}

\date{}

\begin{document}
\maketitle

Advances in artificial intelligence (AI), particularly large language models (LLMs)~\cite{geminiteamgoogleGemini25Pushing2025}, have achieved remarkable success. This progress stems from ``scaling laws'' --- performance improves with greater computation, data, and model size~\cite{kaplanScalingLawsNeural2020}. They now excel at mathematics, medical, legal, and coding exams and competitions.

Yet, this paradigm has a crucial vulnerability: scaling laws are effective only when data is abundant. Human reasoning, which relies on logical operations and abstractions rather than brute-force pattern matching on vast data, proves critical in tackling complex frontier domains, where usable data is often inherently scarce. In such environments, scaling's power diminishes, and LLMs often fail at the complex, multi-step reasoning required for innovation. These challenges do not mean the scaling paradigm has reached its ceiling. Rather, it signals that a fundamental shift in strategy is required --- one that moves beyond scaling alone. The path forward lies in augmenting their powerful statistical intuition with a distinctly human capability: the use of symbols derived from human wisdom as a cognitive technology to structure and simplify complexity, especially for navigating these complex, data-scarce frontiers.

A foundational process for this is the human ability of ``quotienting'': creating a compact symbolic space from a vast problem space, in a process akin to forming a mathematical quotient space via equivalence classes. Selecting these symbols requires deep insight. While this emphasis on symbols echoes early symbolic AI paradigms, its role in the modern landscape is transformed. For instance, abductive learning~\cite{daiBridgingMachineLearning2019,zhouAbductiveLearningBridging2019}, a powerful approach from the pre-scaling law era, also refined symbols but assumed sufficient data was available for the model to learn from. Similarly, early neuro-symbolic methods sought to build rigid, self-explanatory systems or explicitly encode expert knowledge~\cite{hitzlerNeurosymbolicApproachesArtificial2022}. In contrast, the new strategy uses symbols as vessels for compressed human wisdom, creating maps that guide the powerful statistical intuition of LLMs, which is especially critical for tackling complex problems where usable data is by nature sparse.

The power of symbols as cognitive technology is illustrated by the Pirahã people ~\cite{frankNumberCognitiveTechnology2008}. Their language lacks words for exact numbers, not even ``one''. Despite this, they can perform one-to-one matching with visible objects, showing grasp of exact quantity. However, precision vanishes on memory tasks. When replicating quantities of spools shown then hidden, estimates became unreliable for small numbers like five, growing increasingly inaccurate with larger quantities. This reveals number words are not the concept's source, but a symbolic invention --- ``cognitive technology'' for remembering and comparing quantities.

The design of a symbol can further amplify intellect, a story told by the competing notations of calculus. Isaac Newton and Gottfried Wilhelm Leibniz independently developed calculus, but through different symbolic languages. Newton used a dot notation ($\dot{y}$) to represent a rate of change over time. This is compact but conceptually opaque. Leibniz designed the notation $\frac{dy}{dx}$, which frames the derivative as a relationship between infinitesimal changes in $y$ and $x$. This design was a stroke of genius. For a student, the chain rule in modern notation, $(f(g(x)))'=f'(g(x))g'(x)$, is a formula to be memorized. In Leibniz's notation, it becomes an intuitive cancellation: $\frac{dy}{dx}=\frac{dy}{du}\cdot\frac{du}{dx}$. The symbol itself helps you guess the rule. The inverse function rule becomes a simple inversion: $\frac{dx}{dy}=1/\frac{dy}{dx}$. Leibniz's notation created a powerful heuristic for thought, remaining the primary language for calculus today.

This principle finds compelling application in AlphaGeometry series ~\cite{trinhSolvingOlympiadGeometry2024a,chervonyiGoldmedalistPerformanceSolving2025}, a system reaching gold-medal level in International Mathematical Olympiad (IMO). Its success is best understood through the historical evolution of geometric problem-solving. Classical Euclidean geometry relied on human insight but was difficult to automate. The first symbolic leap, Descartes's analytic geometry, translated geometric problems into algebra, a powerful method often leading to intractable search spaces. AlphaGeometry represents a ``helical return'' --- a synthesis that revives the classical framework by augmenting it with a neuro-symbolic engine. An LLM, trained on a human-designed symbolic language for geometric constructions, does not solve the entire problem. It learns to make the crucial human-like leap: proposing an auxiliary line. This action is fed to a deductive solver that efficiently explores consequences. This synthesis overcomes limitations of both prior paradigms.

This synergy of symbolic guidance and statistical intuition is emerging as a powerful paradigm, creating promising open fields for research and discovery:
\begin{itemize}
\item \textbf{Algorithm design with theoretical bounds.} This research frontier aims to automatically discover novel algorithms with provable guarantees. For example, in the \textbf{LegoNE} framework, an LLM guided by a symbolic compiler designed an algorithm for computing approximate Nash equilibria that surpassed all previous methods developed by human experts~\cite{dengComputeraidedApproachApproximate2024}. Can we automatically discover novel algorithms with tight theoretical guarantees for unsolved problems in game theory (e.g., correlated equilibria in stochastic games)?

\item \textbf{Combinatorial optimization.} The \textbf{AutoSAT} system uses this approach to automatically refine heuristics for Boolean satisfiability (SAT), a fundamental NP-complete problem. By directly modifying the symbolic rules within a SAT solver's source code, an LLM guided by experimental outcomes systematically improved its performance, a critical task for domains like chip design~\cite{sunAutoSATAutomaticallyOptimize2024}. How can neuro-symbolic systems dynamically adapt heuristics across diverse SAT problem distributions without human intervention?

\item \textbf{Algorithm design for computer systems.} The paradigm extends to generating highly optimized code for specific hardware. For instance, engineers at \textbf{NVIDIA} used an LLM guided by an iterative verifier to produce GPU kernel code for the attention mechanism that outperformed expertly hand-tuned libraries, demonstrating a powerful new method for systems optimization~\cite{chenAutomatingGPUKernel2025}. Can we develop architecture-aware symbolic abstractions that enable LLMs to optimize entire neural network architectures (not just kernels) for novel hardware?
\end{itemize}

Thus, the next frontier for AI will not be conquered by scaling alone. The art of symbolization itself --- the crafting of powerful abstractions --- is therefore the central task ahead. If scaling laws have given models their powerful intuition, it is the art of the symbol that will provide the compass for genuine discovery.

\paragraph{Acknowledgment.} This work is supported the Natural Science Foundation of China (Grant No. 62172012).

\bibliographystyle{unsrt}
\bibliography{ref}

\begin{thebibliography}{10}

\bibitem{geminiteamgoogleGemini25Pushing2025}
{Gemini Team, Google}.
\newblock Gemini 2.5: {{Pushing}} the {{Frontier}} with {{Advanced Reasoning}}, {{Multimodality}}, {{Long Context}}, and {{Next Generation Agentic Capabilities}}.
\newblock \url{https://storage.googleapis.com/deepmind-media/gemini/gemini_v2_5_report.pdf}, June 2025.
\newblock (accessed 2025-08-14).

\bibitem{kaplanScalingLawsNeural2020}
Jared Kaplan, Sam McCandlish, Tom Henighan, Tom~B. Brown, Benjamin Chess, Rewon Child, Scott Gray, Alec Radford, Jeffrey Wu, and Dario Amodei.
\newblock Scaling {{Laws}} for {{Neural Language Models}}, January 2020.

\bibitem{daiBridgingMachineLearning2019}
Wang-Zhou Dai, Qiuling Xu, Yang Yu, and Zhi-Hua Zhou.
\newblock Bridging machine learning and logical reasoning by abductive learning.
\newblock In {\em Proceedings of the 33rd {{International Conference}} on {{Neural Information Processing Systems}}}, number 253, pages 2815--2826. Curran Associates Inc., Red Hook, NY, USA, December 2019.

\bibitem{zhouAbductiveLearningBridging2019}
Zhi-Hua Zhou.
\newblock Abductive learning: Towards bridging machine learning and logical reasoning.
\newblock {\em Sci. China Inf. Sci.}, 62(7):76101, March 2019.

\bibitem{hitzlerNeurosymbolicApproachesArtificial2022}
Pascal Hitzler, Aaron Eberhart, Monireh Ebrahimi, Md~Kamruzzaman Sarker, and Lu~Zhou.
\newblock Neuro-symbolic approaches in artificial intelligence.
\newblock {\em Natl Sci Rev}, 9(6), June 2022.

\bibitem{frankNumberCognitiveTechnology2008}
Michael~C. Frank, Daniel~L. Everett, Evelina Fedorenko, and Edward Gibson.
\newblock Number as a cognitive technology: {{Evidence}} from {{Pirahã}} language and cognition.
\newblock {\em Cognition}, 108(3):819--824, September 2008.

\bibitem{trinhSolvingOlympiadGeometry2024a}
Trieu~H. Trinh, Yuhuai Wu, Quoc~V. Le, He~He, and Thang Luong.
\newblock Solving olympiad geometry without human demonstrations.
\newblock {\em Nature}, 625(7995):476--482, January 2024.

\bibitem{chervonyiGoldmedalistPerformanceSolving2025}
Yuri Chervonyi, Trieu~H. Trinh, Miroslav Olšák, Xiaomeng Yang, Hoang Nguyen, Marcelo Menegali, Junehyuk Jung, Vikas Verma, Quoc~V. Le, and Thang Luong.
\newblock Gold-medalist {{Performance}} in {{Solving Olympiad Geometry}} with {{AlphaGeometry2}}, February 2025.

\bibitem{dengComputeraidedApproachApproximate2024}
Xiaotie Deng, Dongchen Li, and Hanyu Li.
\newblock A {{Computer-aided Approach}} for {{Approximate Nash Equilibria}}.
\newblock In {\em Web and {{Internet Economics}} - 20th {{International Conference}}, {{WINE}} 2024, {{Edinburgh}}, {{UK}}, {{December}} 2-5, 2024, {{Proceedings}}}, Edinburgh, UK, December 2024.

\bibitem{sunAutoSATAutomaticallyOptimize2024}
Yiwen Sun, Xianyin Zhang, Shiyu Huang, Shaowei Cai, BingZhen Zhang, and Ke~Wei.
\newblock {{AutoSAT}}: {{Automatically Optimize SAT Solvers}} via {{Large Language Models}}, May 2024.

\bibitem{chenAutomatingGPUKernel2025}
Terry Chen, Bing Xu, and Kirthi Devleker.
\newblock Automating {{GPU Kernel Generation}} with {{DeepSeek-R1}} and {{Inference Time Scaling}}.
\newblock \url{https://developer.nvidia.com/blog/automating-gpu-kernel-generation-with-deepseek-r1-and-inference-time-scaling/}, February 2025.
\newblock (accessed 2025-07-08).

\end{thebibliography}
\end{document}